\documentclass[journal]{IEEEtran}

\usepackage{cite}
\usepackage{amssymb}
\usepackage{algorithm}
\usepackage{algorithmic}
\usepackage{epsfig}
\usepackage{subfigure}
\usepackage{multirow}
\usepackage{float}
\usepackage{amsmath}
\usepackage{longtable}
\usepackage{booktabs}
\usepackage{color}
\usepackage{subfigure}
\usepackage{fancyhdr}
\usepackage{epstopdf}
\usepackage{bm}
\usepackage{amsopn}
\usepackage{lineno}
\begin{document}
\title{Posture Adjustment for a Wheel-legged Robotic System via Leg Force Control with Prescribed Transient Performance}
%Dynamic Attitude Adjustment for a Wheel-legged Robot via Impedance Control of Stewart Structure}
\author{Dongchen~Liu,~\IEEEmembership{Student Member,~IEEE},
        Junzheng~Wang,
        Dawei~Shi,~\IEEEmembership{Senior Member,~IEEE},
        Shoukun~Wang,
        Huaihang~Zheng,
        and~Yuan~Huang

\thanks{This paper is submitted for review on March 20, 2021. This work was supported in part by the National Key Research and Development Project of China under Grant 2019YFC1511401 and the National Natural Science Foundation of China under Grant 51675041 and Grant 61773060. (Corresponding author: Junzheng Wang)}
\thanks{D. Liu, J. Wang, D. Shi, S. Wang, and H. Zheng are with the School of Automation, Beijing Institute of Technology, Beijing 100081, China (e-mails: liudongchen@bit.edu.cn, wangjz@bit.edu.cn, daweishi@bit.edu.cn, bitwsk@bit.edu.cn, ss8zzhhhang@163.com). }
\thanks{Y. Huang is with the Science and Technology on Space Intelligent Control Laboratory, Beijing Institute of Control Engineering, Beijing 100094, China (e-mail: yuan$\_$huang@bit.edu.cn).}} % <-this % stops a space


\maketitle
\begin{abstract}

This work proposes a force control strategy with prescribed transient performance for the legs of a wheel-legged robotic system to realize the posture adjustment on uneven roads. A dynamic model of the robotic system is established with the body postures as inputs and the leg forces as outputs, such that the desired forces for the wheel-legs are calculated by the posture reference and feedback. Based on the funnel control scheme, the legs realize force tracking with prescribed transient performance. To improve the robustness of the force control system, an event-based mechanism is designed for the online segment of the funnel function. As a result, the force tracking error of the wheel-leg evolves inside the performance funnel with proved convergence. The absence of Zeno behavior for the event-triggering condition is also guaranteed. The proposed control scheme is applied to the wheel-legged physical prototype for the performance of force tracking and posture adjustment. Multiple comparative experimental results are presented to validate the stability and effectiveness of the proposed methodology.

\end{abstract}


\IEEEpeerreviewmaketitle

\section{Introduction}

Research on wheeled mechanisms has a long tradition due to its advantages of carrying capability, energy efficiency, and motion speed \cite{wheel2016,SY2016,Liu2019Dec,SY2020,Tmech2020,Lin2020Nonlinear}. In addition to the remarkable progress in the simultaneous localization and mapping (SLAM) \cite{IROS2} as well as the path tracking control schemes \cite{TieWM2,PT4,TVTluan}, the development of body stability for wheeled robotic systems on uneven surfaces has attracted increasing attentions \cite{AW3,AW4,AW5,Huang2020Anti}. The high performance of locomotion safety is of primary importance for autonomous missions, e.g., planetary or volcanic exploration, hazardous rescue, and field prospecting. However, the main challenge arises from ensuring body stability, which is the field of study that focuses on balance maintenance and posture dynamic adjustment.

Considerable efforts have been made towards the design and control for active suspension systems, which are responsible for transmitting and filtering all forces between the wheeled system and the road \cite{wu2019design,TieASS,TVTwu}. In \cite{TieASS}, a novel control strategy was presented for nonlinear uncertain vehicle active suspension systems without using any function approximations, e.g., neural networks or fuzzy logic systems. A preview semi-active suspension control was designed in \cite{TVTwu} for autonomous vehicles on uneven roads, such that the ride comfort optimization was realized. In conclusion, the control system achieves suspension response by adding or dissipating energy from the wheeled mechanism with the actuators between the body and the wheels. The kinetic energy was effectively reduced in practice, such that the control systems realized satisfying posture stability. However, the methods include no attitude sampling and control, which limit their practical application for the posture dynamic adjustment.

\begin{figure*}
	\centering
	\includegraphics[width=1.0\hsize]{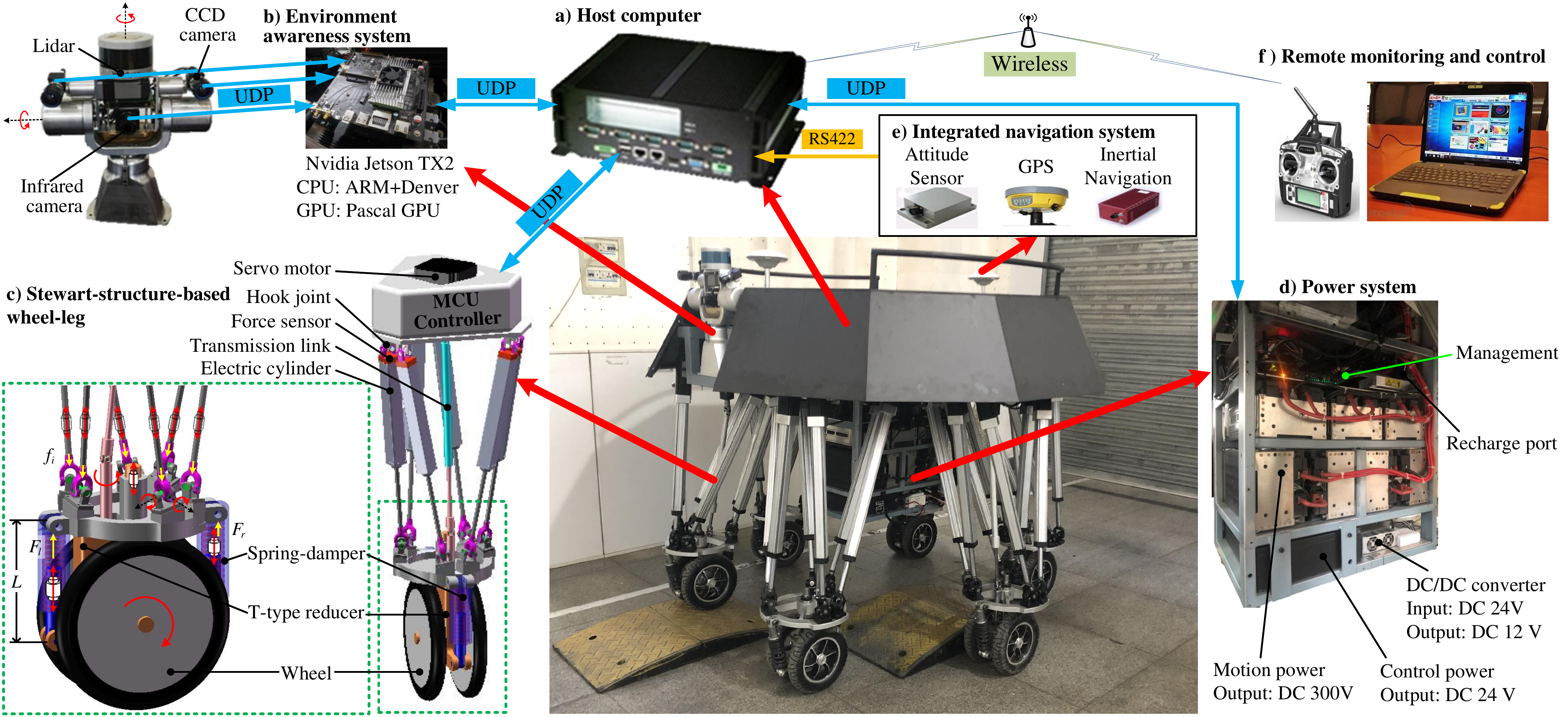}
	\caption{Physical prototype of the wheel-legged robotic system BITNAZA.}
	\label{figRS}
\end{figure*}

The wheels equipped with legs help the robots to realize flexible movement \cite{RW96,WP03,AW1}, such as the wheel driving similar to vehicles, the foot walking similar to legged robots, and compound motion. The wheel-legged robotic system can simultaneously realize wheel driving and posture adjustment in compound motion, which strengthens the terrain adaptability. A wheel-legged robot Hylos \cite{Hylos1} with a velocity-based posture adjustment algorithm was carried out for traveling on irregular sloping terrain. As every movement was planned \cite{Hylos2}, the errors on the ground position invalidated the computed optimization of load distribution due to the stiffness of the structures. Considering the forces at each contact between the wheel and ground, the phenomenon can be alleviated by an adaptive impedance control to govern the robot by the target impedance model \cite{RA17}. Based on virtual model control, the study was fundamentally improved in \cite{Hylos3} by estimating the ground contact angles and then enhancing the posture control \cite{Hylos4}. Pioneered by the works in \cite{VC2001}, numerous results have been obtained in virtual model control to strengthen the locomotion performance \cite{VCTAC,VCTVT,VC2019}. In such a control scheme, a virtual model is introduced with the body posture of the robot as the function of a virtual force. Based on the tracking control, the vector sum of the forces from actuators as well as the legs tracks the virtual force. An optimized virtual model reference control synthesis method was proposed in \cite{VCTVT} based on the ride and vehicle handling characteristics. The aforementioned investigations mainly focused on the asymptotic stability for the virtual model control systems and ignored the transient performance, which was a significant point to improve the posture adjusting performance. Related results, as well as experimental data, are still scarce, which motivates this work.

In this work, the posture control problem for a wheel-legged robotic system is considered to enhance the locomotion stability on an uneven surface. Based on the analysis of the robot dynamics, the desired forces of the wheel-legs are planned for balance maintenance and posture adjustment of the body. To ensure prescribed transient performance for the force tracking of each wheel-leg, the funnel control strategy is adopted, which is originally proposed in \cite{FCo,FCtm,FCsiso,FCmimo} and attracts much attention recently \cite{FCsy,errouissi2018experimental,na2018active}. In this strategy, the gain of the control law is determined dynamically based on the tracking error and a specified performance funnel. As a result, the tracking error evolves within the performance funnel, which is an off-line designed function of time in the existing literature \cite{na2018active}. However, the complexity of the uneven road varies, which is mainly manifested by the amplitude and frequency of ups and downs. When the robotic system is stabilized with the prescribed transient performance, the system may not handle the complexity changes with unstable performances. Therefore, the online segment is necessary for the funnel function design. In this way, the prescribed transient behavior and asymptotic tracking are robustly achieved, while little attention has been focused on this problem. Motivated by the previous research on event-based sampling and control \cite{Event0,Event1,Chen2017Event,Huang}, we make efforts for the event-based segment. The main contributions of our work are as follows:
\begin{itemize}
	\item A performance funnel is designed with an event-based mechanism to realize the online segment. When the control system no longer handles the force tracking, the event-based mechanism is triggered, such that the funnel function is refreshed. The absence of continuous update is guaranteed by the inexistence of the Zeno phenomenon.
	\item To enhance the locomotion stability for the wheeled system on an uneven surface, the posture adjustment method is carried out. The dynamic model of the robotic system is established with the body postures as inputs and the desired forces of legs as outputs. Based on the developed performance funnel, the controllers for wheel-legs are designed to realize force tracking with proved convergence.
	\item The developed posture dynamic adjustment results are evaluated on a wheel-legged robot with legs of Stewart structure. The reliability and performance of the proposed controller are demonstrated with multiple experiments of the robot under simulating unstructured environments.
\end{itemize}

The rest of this paper is organized as follows. In Section II, the system description for a wheel-legged robotic system to be controlled and the problem formulation are presented. The dynamics analyses for the body and the single-leg are proposed. In Section III, the force tracking scheme based on funnel control strategy is shown in detail. The convergency and stability of the force tracking control system are guaranteed with prescribed transient performance. Experimental results are totally described in Section IV, and concluded with a brief discussion in Section V.

%SMCVPR,IROS1,  PT2,PT1,

\section{System Description and Problem Formulation}

\subsection{System Overview}

\begin{figure*}
	\centering
	\includegraphics[width=1.0\hsize]{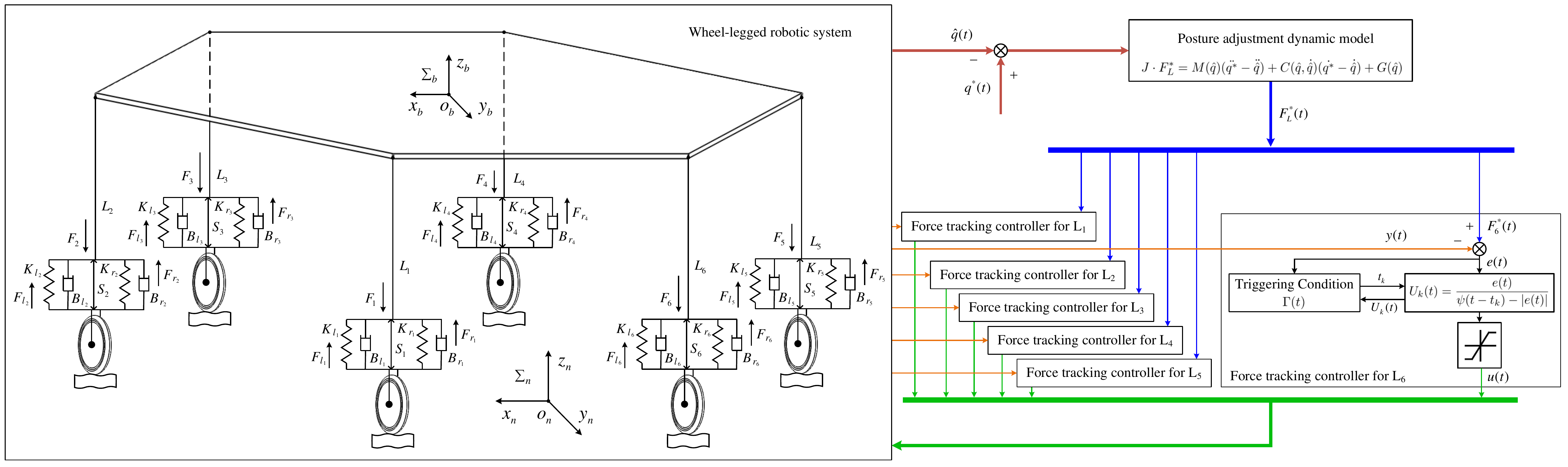}
	\caption{Coordinate system of the wheel-legged robot with dynamic analysis and the block diagram of the posture adjustment strategy.}
	\label{figRD}
\end{figure*}

Aiming at exploring efficient wheeled mechanisms to improve the locomotion performance, we have developed a wheel-legged robot named BITNAZA as shown in Fig.~\ref{figRS} with the structure parameters in Table~\ref{tab:addlabel}. Equipped with the environment awareness system and the integrated navigation system, the all-weather simultaneous localization and mapping (SLAM) have been realized. Based on the SLAM results through the user datagram protocol (UDP) and serial port RS422, the autonomous motion decision and parameter optimization are achieved on the host computer. With the management and converters, the power system provides energy to the robot with a battery life of up to 4 hours. For safety reasons, remote monitoring and control are essential. Driven by 6 electric cylinders in the Stewart structure, the legs realize decoupling motion in 6 degrees-of-freedom (DOF). The wheels are connected with T-type reducers, transmission links, and servo motors to get motion torque, and are suspended on the legs through 2 spring-dampers and a linear bearing. Benefiting from the wheeled legs, the robotic system has strong loading capacity and terrain adaptability. The flexible multi motions are applied to circumvent buildings and cross obstacles by adjusting the supporting polygon.

\begin{table}[!htb]
	\centering
	\caption{Parameters of the Physical Prototype}
	\label{tab1}
	\begin{tabular}{cccc}
		\hline\hline
		Parameter Name & Value \\
		\hline
		Robot Length$*$Width$*$Height (m) & 1.6$*$1.6$*$1.2  \\
		Robot Weight (kg) & 436.0 \\
		Carrying Capacity (kg) & 380.0 \\
		Maximum Angle of climb ($^o$) & 27.0\\
		Maximum velocity (km/h) & 10.0 \\
		Battery Life (h) & 4.0\\
		\hline\hline
	\end{tabular}%
	\label{tab:addlabel}%
\end{table}

In practice, BITNAZA can simultaneously realize wheel driving and leg adjusting in compound motion. The key problem of wheel motion stability focuses on the posture adjustment strategy to strengthen locomotion performance on the complexed ground. This challenge motivates our investigation.

\subsection{Dynamics Modeling}

For the wheel-legged robotic system, the description of the control model is divided into two parts, i.e., the body dynamics and the single-leg dynamics.

As shown in Fig.~\ref{figRD}, coordinate system $\Sigma_b$ and $\Sigma_n$ are established according to the right-hand rule. In $\Sigma_b$, the origin $o_b$ is fixed on the body coinciding with the center of gravity (COG), the $x_b-$axis is the longitudinal axis of the system (the forward direction is positive), and the $y_b-$axis is the lateral axis of the system (the right-to-left direction is positive). In $\Sigma_n$, $x_n-$axis positive direction is due east, and $y_n-$axis positive direction is due north. This model is 6 degrees of freedom (DOF), and $q\in\mathbb{R}^{6}$ represents the translational and rotational positions of $\Sigma_b$ in $\Sigma_n$. The wheel-legs are initially located at the vertices of a regular-hexagonal body with the length $l$, and numbered in the clockwise direction with $L_i$ $(i=1,...,6)$. It is assumed that the mechanical properties of each leg are the same.

Considering the dynamic analysis of the robotic system, a 6-dimensional force $\tau \in \mathbb{R}^{6}$ represents the vector sum of the planned forces $F_i^*$ $(i=1,...,6)$ from wheel-legs to the body, such that
\begin{equation}
\label{eqDM2}
\begin{aligned}
{\tau} = \left[ {\begin{array}{*{20}{c}}
	\sum\limits_{i=1}^6 \left(F_i^* \cdot {\frac{{P_i} - {B_i}}{\left| {{P_i} - {B_i}} \right|}}\right)\\
	\sum\limits_{i=1}^6 \left(F_i^* \cdot {\frac{{P_i} \times ({P_i}-{B_i})}{\left| {{P_i} - {B_i}} \right|}}\right)
	\end{array}} \right].
\end{aligned}
\end{equation}
We write
\begin{equation}
\label{eqDM3}
\begin{aligned}
\tau = J \cdot F_L^*,
\end{aligned}
\end{equation}
where $J$ is the Jacobian matrix of the system and $F_L^*$ is the matrix of the planned force with
\begin{equation}
\label{eqDM4}
\begin{aligned}
J&= \left[ {\begin{array}{*{20}{c}}
	\frac{P_1 - B_1}{\left|P_1 - B_1\right|}& \frac{P_2 - B_2}{\left| P_2 - B_2\right|}&{...}&\frac{P_6 - B_6}{\left| P_6 - B_6 \right|}\\
	\frac{-P_1 \times B_1}{\left|P_1 - B_1\right|}& \frac{-P_2 \times B_2}{\left|P_2 - B_2\right|}&{...}&\frac{-P_6 \times B_6}{\left|P_6 - B_6\right|}
	\end{array}} \right],\\
F_L^*&=\left[ {\begin{array}{*{20}{cccccc}}
	F_1^*& F_2^*& F_3^*& F_4^*& F_5^*& F_6^*
	\end{array}} \right]^T.
\end{aligned}
\end{equation}
Then, the force balance equation is set up as
\begin{equation}
\label{eqDM1}
\begin{aligned}
\tau &= M(q) \ddot q + C(q,\dot q)\dot q + G(q),\\
\end{aligned}
\end{equation}
where $M(q)\in \mathbb{R}^{6 \times 6}$ is generalized mass of the body, $C(q,\dot q)\in \mathbb{R}^{6 \times 6}$ represents Coriolis and centrifugal force, and $G(q)\in \mathbb{R}^{6}$ is gravitational force.

In Fig.~\ref{figRD}, the force analysis for single wheel-leg is also presented in detail. As the mechanical properties of each leg, $B_{r_i}$ and $B_{l_i}$ are the coefficients of damping, $K_{r_i}$ and $K_{l_i}$ are the coefficients of stiffness, and $S_i$ is the length variation of the spring-dampers in wheel-leg $L_i$. We have
\begin{equation}
\label{eqDM5}
\begin{aligned}
F_i&= F_{r_i} + F_{l_i}\\
&= (B_{r_i} \dot{S_i} + K_{r_i} S_i) + (B_{l_i} \dot{S_i} + K_{l_i} S_i)\\
&= B_i \dot{S_i} + K_i S_i,
\end{aligned}
\end{equation}
where $B_i = B_{r_i}+B_{l_i}$ and $K_i = K_{r_i}+K_{l_i}$ with $i=1,...,6$.

\subsection{Problem Formulation}

For the reference $q^*(t)$ and the posture feedback $\hat{q}(t)$, we consider the model in (\ref{eqDM2})-(\ref{eqDM1}), such that it is obtained
\begin{equation}
\label{eqPF1}
\begin{aligned}
J \cdot F_L^* = M(\hat{q}) (\ddot {q^*}-\ddot{\hat{q}})+ C(\hat{q},\dot {\hat{q}})(\dot {q^*}-\dot{\hat{q}}) + G(\hat{q}).
\end{aligned}
\end{equation}
Taking a derivation on both sides of (\ref{eqDM5}), it holds
\begin{equation}
\label{eqPF2}
\begin{aligned}
\dot{F}_i(t) = B_i \ddot{S_i}(t) + K_i \dot{S_i}(t).
\end{aligned}
\end{equation}
Let $x(t) = [x_1(t), x_2(t)]^T = [S_i(t), \dot S_i(t)]^T$ represent the system states. Considering the force tracking problem for single leg, $F_i(t)$ is the output. As the legs are active mechanisms, the input $u(t)$ is designed to be linear related to $\dot{F}_i(t)$ with $k_p$. With the initial value $x_0=x(t_0)=[0,0]^T$, the force tracking model to be controlled is given as
\begin{equation}
\label{eqPF3}
\left\{
\begin{aligned}
\dot x_1(t) &= x_2(t),\\
\dot x_2(t) &= \frac{k_p}{B_i}u(t) - \frac{K_i}{B_i} x_2(t),\\
y(t) &= K_i x_1(t)+ B_i x_2(t).
\end{aligned}
\right.
\end{equation}
For notion brevity, we define $\rho_1= -\frac{K_i}{B_i}$, $\rho_2= \frac{k_p}{B_i}$, $\rho_3=K_i$, and $\rho_4=B_i$. The tracking error is
\begin{equation}
\label{eqPF5}
\begin{aligned}
e(t) = F_i^*-y(t).\\
\end{aligned}
\end{equation}

In this work, we mainly focus on the posture adjustment via force tracking control for the wheel-legs when the robotic system moves on uneven roads. In particular, we mainly study the following problems:
\begin{itemize}
	\item Considering the control system in (\ref{eqPF3}), can we propose a funnel function $\psi(t)$ with an event-based online segment and guarantee the absence of the Zeno phenomenon for the condition $\Gamma(t)$.
	\item Based on the presented funnel function $\psi(t)$ and event-triggering condition $\Gamma(t)$, can we design a control signal $u(t)$, and guarantee the convergence of the system in the sense of $|e(t)| \leq \psi(t)-\varpi$, where $\varpi$ is a positive constant?
\end{itemize}

\section{Workspace based Impedance Control of the Stewart Structure}

In this section, the problems stated above are investigated in theory. With a constructor $z(t)$ and the initial value $z_0=z(t_0)=0$, the control system in (\ref{eqPF3}) is transformed as
\begin{equation}
\label{eqs-1}
\begin{aligned}
\dot x_1(t)&= x_2(t),\\
\dot x_2(t)&=f(x(t),z(t))+\rho_2u(t),\\
\dot z(t)  &=g(x(t),z(t)),\\
y(t)  &=h(x(t),z(t)),
\end{aligned}
\end{equation}
where the functions $f(x(t),z(t))$, $g(x(t),z(t))$, and $h(x(t),z(t))$ satisfy
\begin{equation}
\label{eqs-2}
\begin{aligned}
f(x(t),z(t))&=\rho_1x_2(t),\\
g(x(t),z(t))&=-z(t),\\
h(x(t),z(t))&=\rho_3x_1(t)+\rho_4x_2(t).
\end{aligned}
\end{equation}
Therefore, it is obtained that $\|z(t)\|=0$ holds for $t>t_0$. If we choose
\begin{equation}
\label{eqs-3}
\left\{
\begin{aligned}
\beta(\star,t)&=\frac{\star}{\textrm{exp}(t)},\\
\alpha(\star)&=\star,\\
\end{aligned}
\right.
\end{equation}
$\dot{z}(t)=-z(t)$ is an input-to-state stable system and satisfies
\begin{equation}
\label{eqs-4}
\begin{aligned}
\|z(t)\| \leq \beta(\|z_0\|,t) + \alpha\left(\textrm{sup}_{\tau \in [0,t]}|x(\tau)|\right).
\end{aligned}
\end{equation}
To describe the transient performance of the control system, the funnel is designed as
\begin{equation}
\label{eqs-5}
\begin{aligned}
\mathcal{F}(\psi(t))=\{(t,e(t))|\psi(t)>|e(t)|\},
\end{aligned}
\end{equation}
where $\psi(t):\mathbb{R}\rightarrow [\xi, +\infty)$ is a bounded function based on a determined parameter $\xi$. As the boundary of tracking error $e(t)$, the funnel $\psi(t)$ has a strongly effect on the performance and is provided with positive constants $a>0$ and $b>0$ as
\begin{equation}
\label{eqs-6}
\begin{aligned}
\psi(t)=\frac{a}{\textrm{exp}(b t)}+\xi.
\end{aligned}
\end{equation}

As the block diagram in Fig.~\ref{figRD}, the control input are determined by an event-triggering mechanism. In this case, the obtained control signal is of the form that
\begin{equation}
\label{eqs-7}
\begin{aligned}
u(t)=-\textrm{sat}\left( \frac{e(t)}{\psi(t-t_k)-|e(t)|} \right),
\end{aligned}
\end{equation}
where
\begin{equation}
\label{eqs-8}
\begin{aligned}
\textrm{sat} (\star) = \left\{ \begin{array}{ll}
\textrm{sgn}(\star)\cdot \hat{U},  &|\star|>\hat{U},\\
\star,             &|\star|\leq \hat{U},
\end{array} \right.
\end{aligned}
\end{equation}
with a positive constant $\hat{U}$ and the symbolic function
\begin{equation}
\label{eqs-9}
\begin{aligned}
\textrm{sgn} (\star) = \left\{ \begin{array}{ll}
1,  &\star>0,\\
-1, &\star<0.
\end{array} \right.
\end{aligned}
\end{equation}
The previous time instant $t_k$ comes from the event-triggering condition. For notion brevity, we write
\begin{equation}
\label{eqs-10}
\left\{
\begin{aligned}
U_k(t) &= \frac{e(t)}{\psi(t-t_k)-|e(t)|},\\
U_s(t) &= \frac{e(t)}{\psi(0)-|e(t)|},\\
\end{aligned}
\right.
\end{equation}
thus, the event-triggering condition is designed as
\begin{equation}
\label{eqs-11}
\begin{aligned}
\Gamma (t) = \left\{ \begin{array}{ll}
0, &\textrm{if}~|U_k(t)-U_s(t)|<\rho_5 \bar{U}(t),\\
1, &\textrm{otherwise},\\
\end{array} \right.
\end{aligned}
\end{equation}
where $\rho_5$ and $\bar{U}(t)$ are designed condition parameters. With a constant $\varrho \in(0,1)$, $\bar{U}(t)$ satisfies
\begin{equation}
\label{eqs-12}
\begin{aligned}
\bar{U}(t) = \left\{ \begin{array}{ll}
\min\{U_k(t),\hat{U}\}, &\textrm{if}~|U_k(t)| > \varrho \hat{U},\\
\varrho \hat{U}, &\textrm{otherwise}.\\
\end{array} \right.
\end{aligned}
\end{equation}
There exists
\begin{equation}
\label{eqs-13}
\left\{
\begin{aligned}
B_f &= \sup_{|x|<B_x} |f(x(t),z(t))|,\\
\Omega &=\|\dot{F}_L^*(t)\|_\infty +\rho_6 B_f+ L_{\psi},
\end{aligned}
\right.
\end{equation}
where $\rho_6 =\frac{\rho_3+\rho_4\rho_1}{\rho_1}$, and $B_x = |x_0| + \|F_L^*\|_\infty + \|\psi\|_\infty $. To guarantee the boundedness of the tracking error $e(t)$, we present the following theorem.

\textit{Theorem 1:} Consider the system in (\ref{eqs-1}), the control law in (\ref{eqs-7}), and the event-triggering condition in (\ref{eqs-11}). If $\hat{U} > \Omega$, $|e(0)|<\psi(0)$, and
\begin{equation}
\label{eqt2-1}
\begin{aligned}
\rho_5 < 1-\frac{\Omega}{\hat{U}},~~ \varrho \hat{U}<1,
\end{aligned}
\end{equation}
there exists $\varpi>0$, such that
\begin{equation}
\label{eqt2-2}
\begin{aligned}
|e(t)| \leq \psi(t)-\varpi, ~~\forall t\in [0,\infty)
\end{aligned}
\end{equation}

\textit{Proof:} According to the control law in (\ref{eqs-7}) and the event-triggering condition in (\ref{eqs-11}), the solution of $\dot{z}(t)=-z(t)$ is represented by $z(t)=\varphi(t,z(t_0),x(t))$. Therefore, the control system in (\ref{eqs-1}) is transformed as
\begin{equation}
\label{eqp2-1}
\begin{aligned}
\!\!\dot{x}_2(t)\!=\!f(x(t),\varphi(t,z_0,x))\!-\!\rho_2\textrm{sat}\left( \frac{e(t)}{\psi(t\!-\!t_k)\!-\!|e(t)|} \right)\!.\\
\end{aligned}
\end{equation}
Considering $e(t)$ in (\ref{eqPF5}), there exists $\kappa\in(0, +\infty)$, such that it holds $|f(x(t),\varphi(t,z_0,x))|<B_f$ for $t\in [0,\kappa)$. Based on (\ref{eqs-10}), it holds
\begin{equation}
\label{eqp2-2}
\begin{aligned}
e(t)\dot{e}(t) &\leq |e(t)|(\Omega-L_{\psi})+\rho_2e(t)u(t)\\
&\leq |e(t)|(\Omega-L_{\psi})-\rho_2e(t)\textrm{sat}(U_k(t)).
\end{aligned}
\end{equation}
For $e(t)>0$, it is obtained
\begin{equation}
\label{eqp2-3}
\begin{aligned}
e(t)\dot{e}(t) \leq |e(t)| (\Omega-\rho_2\textrm{sat}(U_k(t)))-L_{\psi}|e(t)|.
\end{aligned}
\end{equation}
Based on the definition of $\Gamma(t)$ in (\ref{eqs-11}), when $U_k(t) \geq 2\hat{U}$ we have
\begin{equation}
\label{eqp2-4}
\begin{aligned}
\textrm{sat}\left(U_k(t)\right)= \hat{U} \geq \Omega.
\end{aligned}
\end{equation}
When $\hat{U} \leq U_k(t) < 2\hat{U}$, the condition $\Gamma(t)$ in (\ref{eqs-11}) indicates $|U_k(t)-U_s(t)|<\rho_5 \bar{U}(t)= \rho_5 \hat{U}$, such that we observe
\begin{equation}
\label{eqp2-5}
\begin{aligned}
\textrm{sat}(U_k(t))&= \textrm{sat}(U_s(t)+U_k(t)-U_s(t))\\
&\geq(1-\rho_5)\hat{U}.
\end{aligned}
\end{equation}
Due to $\rho_5 < 1-\frac{\Omega}{\hat{U}}$, we have
\begin{equation}
\label{eqp2-6}
\begin{aligned}
\textrm{sat}(U_k(t)) \geq \Omega.
\end{aligned}
\end{equation}
When $\varrho\hat{U} \leq U_k(t) < \hat{U}$, Recalling $\Gamma(t)$ in (\ref{eqs-11}), it is shown that
\begin{equation}
\label{eqp2-7}
\begin{aligned}
|U_k(t)-U_s(t)|<\rho_5 \bar{U}(t) = \rho_5 U_k(t)
\end{aligned}
\end{equation}
holds such that
\begin{equation}
\label{eqp2-8}
\begin{aligned}
\textrm{sat}(U_k(t))&= \textrm{sat}(U_s(t)+U_k(t)-U_s(t))\\
&\geq(1-\rho_5)U_k(t).
\end{aligned}
\end{equation}
It is assumed that for some $t_1 \in [0,\lambda)$
\begin{equation}
\label{eqp2-9}
\begin{aligned}
|e(t_1)|>\psi(t_1)-\varpi,
\end{aligned}
\end{equation}
where the positive constant $\varpi$ is chosen to satisfy
\begin{equation}
\label{eqp2-10}
\begin{aligned}
\varpi\leq \min \left\{ \frac{\xi}{2}, \frac{(1-\rho_5)\xi}{2\Omega}, \psi(0)-|e(0)|\right\}.
\end{aligned}
\end{equation}
According to (\ref{eqp2-9}), $t_0$ is provided as
\begin{equation}
\label{eqp2-11}
\begin{aligned}
t_0=\max \{ t|\psi(t)-|e(t)|=\varpi\}, t\in [0,t_1).
\end{aligned}
\end{equation}
Based on (\ref{eqp2-10}), we have
\begin{equation}
\label{eqp2-12}
\begin{aligned}
\psi(0)-|e(0)|\geq\varpi.
\end{aligned}
\end{equation}
For $t\in [t_0,t_1]$, we obtain
\begin{equation}
\label{eqp2-13}
\begin{aligned}
|e(t)| \geq \psi(t)-\varpi \geq \frac{\xi}{2}.
\end{aligned}
\end{equation}
From the condition $\Gamma(t)$ in (\ref{eqs-11}), it holds
\begin{equation}
\label{eqp2-14}
\begin{aligned}
(1-\rho_5)U_k(t) \geq (1-\rho_5)\frac{\xi}{2\varpi} \geq \Omega.
\end{aligned}
\end{equation}
Relating the results in (\ref{eqp2-4}), (\ref{eqp2-6}), and (\ref{eqp2-14}), we have
\begin{equation}
\label{eqp2-15}
\begin{aligned}
\textrm{sat}(U_s(t)+U_k(t)-U_s(t)) \geq \Omega,
\end{aligned}
\end{equation}
and
\begin{equation}
\label{eqp2-15}
\begin{aligned}
e(t)\dot{e}(t) \leq -L_{\psi}|e(t)|.
\end{aligned}
\end{equation}
Through a simple calculation, it is obtained
\begin{equation}
\label{eqp2-16}
\begin{aligned}
|e(t_1)|-|{e}(t_0)| \leq -L_{\psi}|t_0-t_1|.
\end{aligned}
\end{equation}
Combining with the $\psi(t)$ in (\ref{eqs-6}), we write
\begin{equation}
\label{eqp2-17}
\begin{aligned}
|e(t_1)|-|{e}(t_0)| & \leq -|\psi(t_1)-\psi(t_0)|\\
& \leq \psi(t_1)-\psi(t_0).
\end{aligned}
\end{equation}
Additionally, we have
\begin{equation}
\label{eqp2-18}
\begin{aligned}
\psi(t_1) - |e(t_1)| \geq \psi(t_0) - |{e}(t_0)| = \varpi,
\end{aligned}
\end{equation}
which contradicts the assumption in (\ref{eqp2-9}). As a result, we obtain
\begin{equation}
\label{eqp2-19}
\begin{aligned}
|e(t)| \leq \psi(t)-\varpi.
\end{aligned}
\end{equation}
When $0\leq U_k(t) < \varrho \hat{U}$, it is shown that
\begin{equation}
\label{eqp2-20}
\begin{aligned}
|e(t)| < \frac{\varrho \hat{U}}{1+\varrho \hat{U}} \psi(t).
\end{aligned}
\end{equation}
Considering $\varrho \hat{U}<1$ and $\varpi \leq \frac{\xi}{2}$ from (\ref{eqp2-10}), we have
\begin{equation}
\label{eqp2-21}
\begin{aligned}
\varrho \hat{U} \leq \frac{\psi(t)}{\varpi}-1,
\end{aligned}
\end{equation}
which is equal to (\ref{eqp2-19}). In the case $e(t) < 0$, the convergence of the controller can be guaranteed by a similar analysis. This completes the proof. \hfill $\Box$

\begin{figure*}
	\centering
	\includegraphics[width=1.0\hsize]{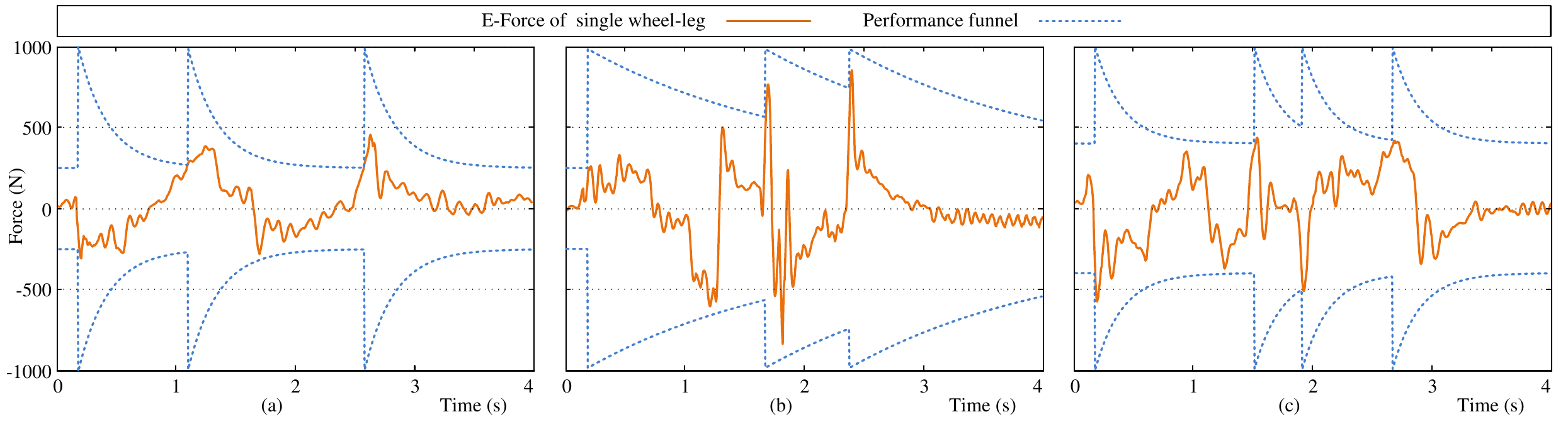}
	\caption{Comparative experiment results of the force tracking performances for single wheel-leg with different parameters: (a) $[a,b,\xi]=[750.0,7.20,250.0]$ in Test 1, (b) $[a,b,\xi]=[750.0,1.25,250.0]$ in Test 2, (c) $[a,b,\xi]=[600.0,7.20,400.0]$ in Test 3.}
	\label{figForce}
\end{figure*}

Note that Theorem 1 and its proof guarantee the tracking performance of the proposed control scheme in the sense that the tracking error $e(t)$ lies in a prescribed range $\psi(t)$ with the control law in (\ref{eqs-7}) and the event-triggering condition in (\ref{eqs-11}). The results provide a practical criterion to plan the transient performance by designing the funnel function $\psi(t)$. In this event-triggering control scheme, the nonexistence of the Zeno phenomenon needs to be proved in the next theorem.

\textit{Theorem 2:} Consider the system in (\ref{eqs-1}), the control law in (\ref{eqs-7}), and the event-triggering condition in (\ref{eqs-11}). There exists $\delta > 0$ such that
\begin{equation}
\label{eqt1-1}
\begin{aligned}
t_{k+1} - t_k \geq \delta
\end{aligned}
\end{equation}
holds for any $k>0$.

\textit{Proof:} Considering the definition in (\ref{eqs-10}), the difference between $U_s(t)$ and $U_k(t)$ can be expressed by
\begin{equation}
\label{eqp1-1}
\begin{aligned}
|U_s(t)-U_k(t)| &= \left|\int_{t_k}^t \frac{d}{d\tau}\left( \frac{e(t)}{\psi(t-\tau) -|e(t)|} \right) d\tau \right|\\
&= \left|\int_{t_k}^t \frac{e(t)\psi(t-\tau)}{\left(\psi(t-\tau)-|e(t)|\right)^2} d\tau \right|\\
&\leq \int_{t_k}^t \frac{|e(t)|\psi(t-\tau)}{\left(\psi(t-\tau)-|e(t)|\right)^2} d\tau\\
\end{aligned}
\end{equation}
According to Theorem 1, it is shown that
\begin{equation}
\label{eqp1-2}
\begin{aligned}
|e(t)| \leq \psi(t)-\varpi.
\end{aligned}
\end{equation}
Combining with the definition of $\psi(t)$ in (\ref{eqs-6}) we have
\begin{equation}
\label{eqp1-3}
\begin{aligned}
|U_s(t)-U_k(t)| &\leq \int_{t_k}^t \frac{(a+\xi)(a+\xi-\varpi)}{\varpi^2} d\tau\\
&\leq\frac{(a+\xi)(a+\xi-\varpi)}{\varpi^2}(t-t_k).
\end{aligned}
\end{equation}
At the time $t=t_{k+1}$, the event-triggering condition holds $\Gamma(t)=1$, which means
\begin{equation}
\label{eqp1-4}
\begin{aligned}
|U_s(t)-U_k(t)| \geq \rho_5 \bar{U}(t).
\end{aligned}
\end{equation}
Therefore, we obtain
\begin{equation}
\label{eqp1-5}
\begin{aligned}
\frac{(a+\xi)(a+\xi-\varpi)}{\varpi^2}(t_{k+1}-t_k) \geq \rho_5 \bar{U}(t).
\end{aligned}
\end{equation}
From the expression of $\bar{U}(t)$ in (\ref{eqs-12}), $\delta$ is designed by
\begin{equation}
\label{eqp1-6}
\begin{aligned}
\delta = \frac{\rho_5 \varrho \hat{U} \varpi^2}{(a+\xi)(a+\xi-\varpi)} > 0,
\end{aligned}
\end{equation}
such that
\begin{equation}
\label{eqp1-7}
\begin{aligned}
t_{k+1} - t_k \geq \delta
\end{aligned}
\end{equation}
holds for any $k>0$. This completes the proof. \hfill $\Box$

Note that Theorem 2 guarantees the existence of the solution in Theorem 1 when $t \rightarrow \infty$. From (\ref{eqs-10}), we have
\begin{equation}
\label{eqp1-8}
\begin{aligned}
U_k(t)-U_s(t) &= \frac{e(t)}{\psi(t-t_k)-|e(t)|}-\frac{e(t)}{\psi(0)-|e(t)|}\\
&= \frac{\psi(0)-\psi(t-t_k)}{\psi(0)-|e(t)|}U_k(t).
\end{aligned}
\end{equation}
Combining the condition $\Gamma(t)$ in (\ref{eqs-11}) with the definition in (\ref{eqs-12}), when $\bar{U}(t) = U_k(t)$, it is obtained for $\Gamma(t)=0$
\begin{equation}
\label{eqp1-9}
\begin{aligned}
\left|\frac{\psi(0)-\psi(t-t_k)}{\psi(0)-|e(t)|}\right|<\rho_5,
\end{aligned}
\end{equation}
such that an explicit relationship between $\rho_5$ and the distance from $|e(t)|$ to $\psi(t-t_k)$ is provided.

\section{Experiment}

In this paper, a series of experiments are carried out on the physical prototype in Fig.~\ref{figRS}. The road with slopes simulates the uneven surface in the real environment.

\subsection{Force Tracking Performance for Single Wheel-leg}

\begin{figure*}
	\centering
	\includegraphics[width=1.0\hsize]{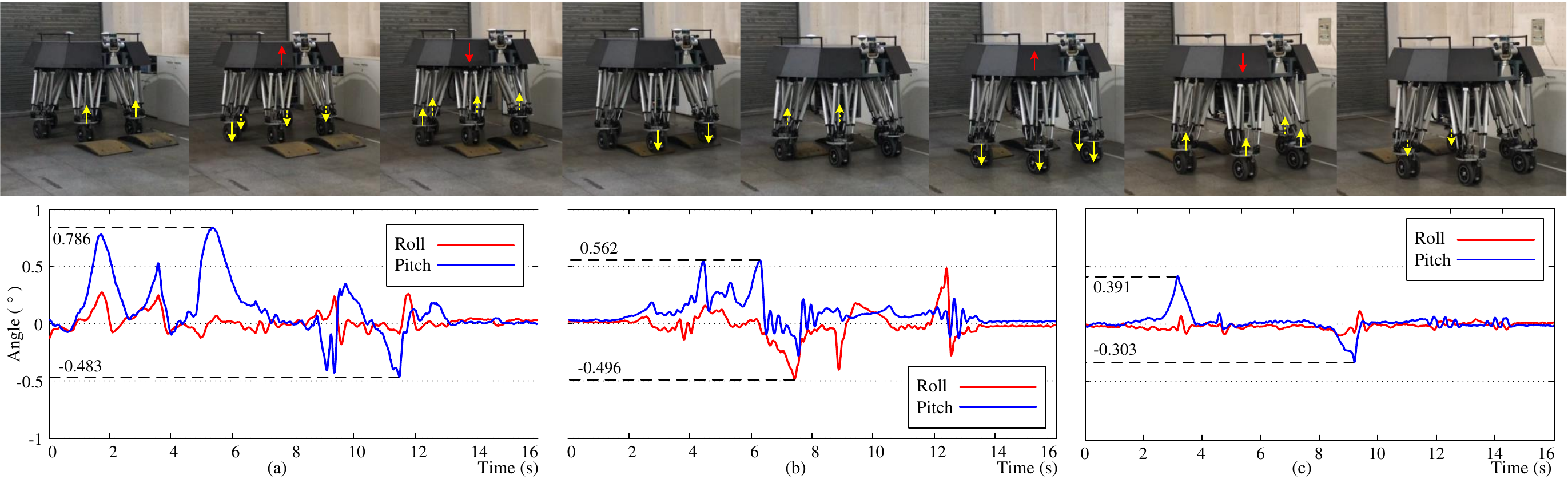}
	\caption{Experiment results on uneven surface with parallel placed slopes: (a) IC with $[m,c,k]=[114.0, 62.5, 109.0]$ for Test 4, (b) FC with $[a,b,\xi]=[600.0,7.20,400.0]$ for Test 5, (c) FC with $[a,b,\xi]=[750.0,7.20,250.0]$ for Test 6.}
	\label{figDoPa}
\end{figure*}

\begin{figure*}
	\centering
	\includegraphics[width=1.0\hsize]{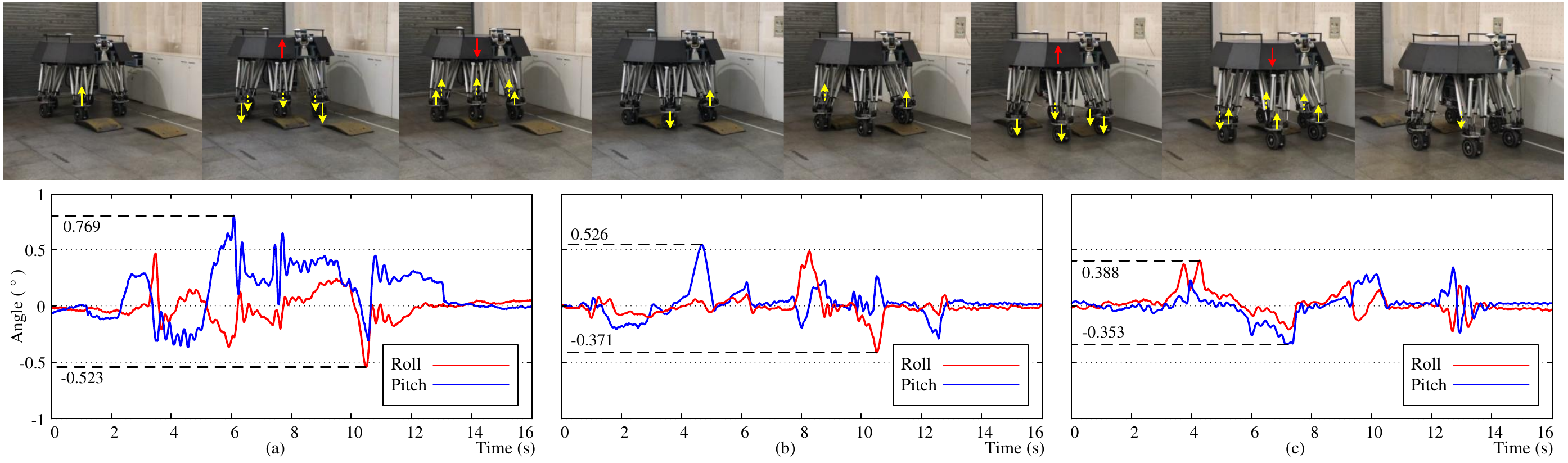}
	\caption{Experiment results on uneven surface with separately located slopes: (a) IC with $[m,c,k]=[114.0, 62.5, 109.0]$ for Test 7, (b) FC with $[a,b,\xi]=[600.0,7.20,400.0]$ for Test 8, (c) FC with $[a,b,\xi]=[750.0,7.20,250.0]$ for Test 9.}
	\label{figDoSe}
\end{figure*}

Comparative experiments for the force tracking performance with different parameters of single wheel-leg are firstly carried out. Based on the prior knowledge, it is chosen that
\begin{equation}
\label{eqE-1}
\begin{aligned}
\psi(0)=a+\xi=1000~(N).
\end{aligned}
\end{equation}
Therefore, the parameters $[a, b, \xi]$ are designed in Table~\ref{tab:para}. The force tracking errors (E-Force) are calculated and shown in Fig.~\ref{figForce}. After the measurement, the E-Force fluctuations are presented in Table~\ref{tab:para}.

\begin{table}[!htb]
	\centering
	\caption{Experiment Parameters and Results of Test 1 to Test 3}
	\begin{tabular}{cccc}
		\hline\hline
		Name      & Parameters & E-Force (N) & Range (N)\\
		\hline
		Test 1    & [750.0, 7.20, 250.0]  & -281.2 $\sim$ 446.3 &  727.5\\
		Test 2    & [750.0, 1.25, 250.0]  & -853.7 $\sim$ 882.6 &  1736.3\\
		Test 3    & [600.0, 7.20, 400.0]  & -543.3 $\sim$ 462.5 &  1005.8\\
		\hline\hline
	\end{tabular}
	\label{tab:para}
\end{table}

With the decrement of $b$, the range of E-Force is reduced from 727.5 N to 1736.3 N. Because the performance funnel $\psi(t)$ changes sharply with $b =\textrm{7.20}$, the E-Force can quickly converge to the expected range in Test 1. Comparing the results for Test 1 and Test 3, the E-Force changement is enlarged as $\xi$ rises, and the reason comes from the performance funnel $\psi(t)$ in Test 3 allows a bigger range for E-Force than that in Test 1. Therefore, the force tracking control of wheel-legs with prescribed transient performance is realized by the proposed funnel control scheme. The results of Test 1 to Test 3 also provide a practical way to select the funnel parameter $[a,b,\xi]$ to achieve satisfied transient performance for force tracking of a single wheel-leg.

\subsection{Posture Adjustment Results}

On the simulated uneven road, the posture adjustment experiments with different control strategies are presented. The roll and pitch angles of the body are used to evaluate the performance. The angles are measured by the MTi-300 series of motion measurement sensors with a sampling frequency of 20 Hz and a baud rate of 115200 bit/s. Two slopes with heights of 50.0 mm and 110.0 mm are parallel placed in Test 4 to Test 6, and separated one behind the other in Test 7 to Test 9. The proposed control scheme (FC) is applied with different performance funnels, and an impedance control (IC) with preset foot trajectory (PFT) is presented for comparison. In reality, the preset foot trajectory can be planned based on the robot vision information, which is the research field of SLAM. For $L_i$ leg, the IC is designed based on a spring-damper-mass model in S-domain as
\begin{equation}
\label{eqE-2}
\begin{aligned}
u_i=\frac{F_i}{ms^2+cs+k},
\end{aligned}
\end{equation}
where $m$, $c$, and $k$ are positive constants, $u_i$ is the displacement input in vertical direction of the leg. Through parameter modification, $[m,c,k]=[114.0, 62.5, 109.0]$ are chosen. Based on the force tracking results for single wheel-leg, $[600.0,7.20,400.0]$ and $[750.0,7.20,250.0]$ is chosen for the $[a,b,\xi]$. The experiment results are presented in Fig.~\ref{figDoPa} and Fig.~\ref{figDoSe}.

\begin{table}[t]
	\centering
	\caption{Experiment Results of Test 4 to Test 6}
	\begin{tabular}{cccc}
		\hline\hline
		Test Number           &Test 4               & Test 5            & Test 6            \\
		\hline
		Controller            &IC                   & FC                &  FC              \\
		PFT                   & Y                   & N                 & N                 \\
		Slops                 & Parallel            & Parallel          & Parallel          \\
		Angle Variation ($^\circ$) & -0.48$\sim$0.79 & -0.50$\sim$0.56 & -0.30$\sim$0.39    \\
		Angle Range ($^\circ$)& 1.27                & 1.06              & 0.69              \\
		\hline\hline
	\end{tabular}
	\label{tab:addlabe2}
\end{table}

\begin{table}[t]
	\centering
	\caption{Experiment Results of Test 7 to Test 9}
	\begin{tabular}{cccc}
		\hline\hline
		Test Number           & Test 7            & Test 8          & Test 9 \\
		\hline
		Controller            &IC                 & FC              & FC\\
		PFT                   & Y                 & N               & N  \\
		Slops                 & Separated         & Separated       & Separated \\
		Angle Variation ($^\circ$)  & -0.52$\sim$0.77   & -0.37$\sim$0.53 & -0.35$\sim$0.39 \\
		Angle Range ($^\circ$)& 1.29              & 0.90            & 0.74 \\
		\hline\hline
	\end{tabular}
	\label{tab:addlabe3}
\end{table}

A series of pictures in Fig.~\ref{figDoPa} record the experiment process of Test 6 with slopes parallel placed, which is similar to Test 4 and Test 5. Continuous multiple pictures in Fig.~\ref{figDoSe} record the experiment process of Test 9 with slopes separately located, which is similar to Test 7 and Test 8. Each wheel-leg of the robotic system extends 80 mm in the vertical direction as the working position. Therefore, the body needs to be raised to pass through uneven roads after retracting its legs to the limit of the workspace. The arrows on the body show the direction of the height adjustment, while the arrows on the wheel-legs represent the directions of relative movements to the body. To illustrate the efficiency of the proposed algorithm, Table~\ref{tab:addlabe2} and Table~\ref{tab:addlabe3} are presented. For the preset foot trajectory (PFT), N means inexistence while Y means existence.

From the comparison, the posture angles in the tests with FC fluent in a smaller range than those in the tests with IC. Meanwhile, the robotic system controlled by FC has satisfied performance in the tests with $ [a,b,\xi]=[750.0,7.20,250.0] $, which is consistent with the results of force tracking for a single wheel-leg. Considering the existence of PFT, the posture adjustment with IC puts forward additional requirements of robot vision for mapping, while the robot with FC realizes blind uneven surface crossing.

The proposed control scheme produces a novel kind of compliance resulting from flexibility in the posture. With the whole control architecture, the robot moves on uneven roads without the need for prior knowledge on their geometry or robot vision data while ensuring certain stability.

\section{Discussion}

This paper studies a posture adjustment strategy for a wheel-legged robotic system through the force tracking control of the wheel-legs with prescribed transient performance. The dynamic model of the robotic system is established with the body postures as inputs and the desired forces of legs as outputs. The funnel control is adopted for the wheel-legs to realize force tracking with prescribed transient performance. As a result, the body posture of the robotic system is adjusted to strengthen the locomotion stability. The experiment results verify the effectiveness of the proposed algorithm.

\bibliographystyle{IEEEtran}
\bibliography{reference}

\vfill


\end{document}